\documentclass[conference]{IEEEtran}
\IEEEoverridecommandlockouts

\usepackage{graphicx}
\usepackage{amsmath}
\usepackage{amssymb}
\usepackage{booktabs}
\usepackage[accsupp]{axessibility} 
\usepackage[pagebackref,breaklinks,colorlinks]{hyperref}

\usepackage[capitalize]{cleveref}
\crefname{section}{Sec.}{Secs.}
\Crefname{section}{Section}{Sections}
\Crefname{table}{Table}{Tables}
\crefname{table}{Tab.}{Tabs.}

\def\wacvPaperID{983} 
\def\confName{WACV}
\def\confYear{2024}

\def\BibTeX{{\rm B\kern-.05em{\sc i\kern-.025em b}\kern-.08em
    T\kern-.1667em\lower.7ex\hbox{E}\kern-.125emX}}

\begin{document}

\title{Residual Graph Convolutional Network for Bird’'s-Eye-View Semantic Segmentation}

\author{\IEEEauthorblockN{1\textsuperscript{st} Qiuxiao Chen}
\IEEEauthorblockA{\textit{Computer Science} \\
\textit{Utah State University}\\
Old Main Hill Logan, Utah \\
chenqiuxiao.ee@gmail.com}
\and
\IEEEauthorblockN{2\textsuperscript{nd} Xiaojun Qi}
\IEEEauthorblockA{\textit{Computer Science} \\
\textit{Utah State University}\\
Old Main Hill Logan, Utah \\
xiaojun.qi@usu.edu}
}
\maketitle

\begin{abstract}
   Retrieving spatial information and understanding the semantic information of the surroundings are important for Bird’s-Eye-View (BEV) semantic segmentation. In the application of autonomous driving, autonomous vehicles need to be aware of their surroundings to drive safely. However, current BEV semantic segmentation techniques, deep Convolutional Neural Networks (CNNs) and transformers, have difficulties in obtaining the global semantic relationships of the surroundings at the early layers of the network. In this paper, we propose to incorporate a novel Residual Graph Convolutional (RGC) module in deep CNNs to acquire both the global information and the region-level semantic relationship in the multi-view image domain. Specifically, the RGC module employs a non-overlapping graph space projection to efficiently project the complete BEV information into graph space. It then builds interconnected spatial and channel graphs to extract spatial information between each node and channel information within each node (i.e., extract contextual relationships of the global features).  Furthermore, it uses a downsample residual process to enhance the coordinate feature reuse to maintain the global information.  The segmentation data augmentation and alignment module helps to simultaneously augment and align BEV features and ground truth to geometrically preserve their alignment to achieve better segmentation results. Our experimental results on the nuScenes benchmark dataset demonstrate that the RGC network outperforms four state-of-the-art networks and its four variants in terms of IoU and mIoU. The proposed RGC network achieves a higher mIoU of 3.1\% than the best state-of-the-art network, BEVFusion. Code and models will be released.
\end{abstract}

\section{Introduction}
\label{sec:intro}

Semantic segmentation of Bird’s-Eye-View (BEV) is a considerable challenge 
task in applications including autonomous driving technology, robotics, autonomous warehouse and factory systems, and so on. For example, autonomous vehicles need to be aware of their surroundings when navigating. To drive safely, autonomous vehicles must be able to 
retrieve spatial information from their environment and understand the semantic information of their surroundings.

Deep convolutional neural networks (CNNs)  \cite{LSS, zhang2022beverse, xie2022m2bev, liu2022bevfusion} and transformers \cite{cvt, li2022bevformer, xu2022cobevt, pan2020cross} are the common backbones of BEV semantic segmentation. Specifically, CNN-based models utilize CNN to extract multi-view image features and process transformed BEV features. However, CNN-based models require plenty of convolutional layers to gain semantic relationships of the whole BEV features due to the limited receptive field of each convolutional layer. On the other hand, transformer-based models concentrate on local patches or part of the global information rather than the overall region relationship when they analyze multi-view image features. CoBEVT\cite{xu2022cobevt} is a transformer-based network that uses sparsely sampled attention to obtain global information. However, it cannot obtain global semantic relationships using the sparsely sampled attention mechanism.

In general, traditional BEV semantic segmentation does not consider contextual relationships of the global features at the early layers of the network. This will lead to inaccurate segmentation, especially at the border of the objects and the small and delicate regions. To address this weakness, we propose to incorporate a novel Residual Graph Convolutional (RGC) module in deep CNNs, to enable BEV semantic segmentation networks to acquire not only the global information but also the region relationship in the BEV map domain produced by multi-view images. Unlike the traditional Graph Convolutional Network \cite{kipf2016semi}, our proposed RGC network utilizes a downsample residual process to enhance the coordinate feature reuse to maintain global information. It also uses a non-overlapping graph space projection to project the complete BEV information into graph space, where interconnected spatial and channel graphs acquire spatial information between each node and channel information within each node to capture the contextual relationships of the global features.

As shown in Fig. 1, a simplified schematic diagram of the proposed RGC network focuses on the two interconnected graphs. The RGC module extracts node features from BEV features to describe relationships between regions in terms of both spatial and channel dimensions before processing BEV features. We build two completely linked graphs, namely the spatial graph and the channel graph, in the RGC module. The spatial graph learns the relationship between each node, where a node represents multi-channel node features and an edge represents the relationships between node features. The channel graph learns the interdependency between each channel, where a node represents node features along one channel and an edge represents the relationships between channels of each node feature. Multi-channel node features of the spatial graph together with the node adjacency matrix and the channel relationship weight of the channel graph within each node are transformed to obtain interactive node features, which are then transformed back into the coordinate space to maintain not only the coordinate information but also the contextual relationship information. As a result, our RGC module efficiently estimates global contextual relationships. Extensive experimental results on the nuScenes dataset \cite{caesar2020nuscenes} show the proposed RGC-based BEV segmentation method achieves state-of-the-art segmentation results of 59.7\% mean IoU. Our contributions are as follows:
\begin{enumerate}
    \item Proposing a novel RGC module consisting of two interconnected graphs (i.e., the spatial graph and the channel graph), where the spatial graph extracts spatial information between each node and the channel graph extracts channel information within each node. The RGC module employs a downsample residual process to enhance the coordinate feature reuse to maintain the global information.  It also employs a non-overlapping graph space projection to efficiently project the complete BEV information into graph space.
    
    \item Incorporating the RGC module that contains both graph and coordinate information in the deep CNNs to enable BEV semantic segmentation networks to not only effectively acquire the global information but also efficiently estimate contextual relationships of the global information in the BEV map domain produced by multi-view images. 
   
    \item Proposing a segmentation data augmentation and alignment module to inhibit the overfitting and misalignment problem in multi-view image and BEV domains by simultaneously augmenting and aligning BEV features and ground truth to geometrically preserve their semantic alignment.
\end{enumerate}

\begin{figure}[t]
 \includegraphics[width=0.95\linewidth]{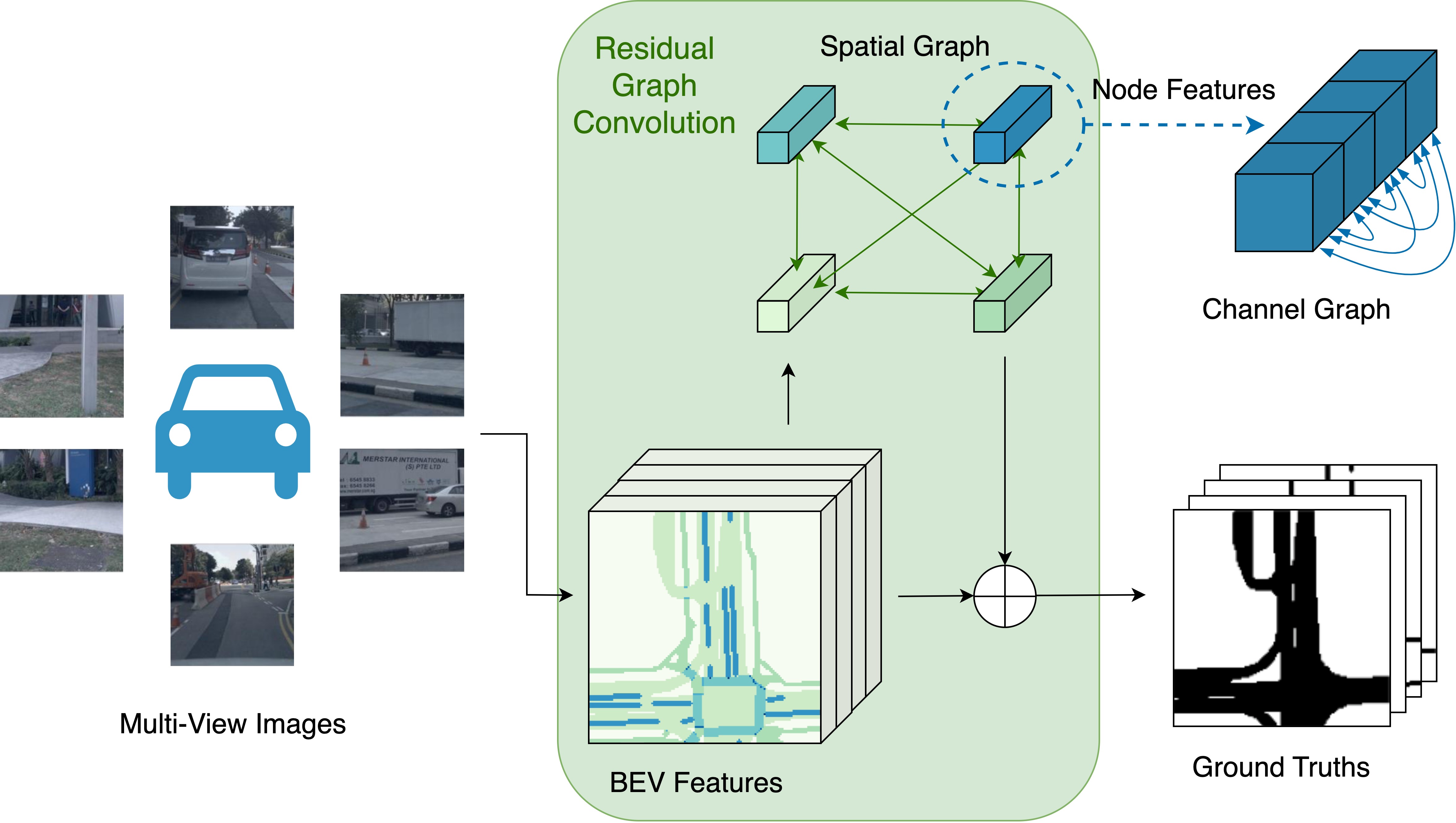}
  \caption{A simplified schematic diagram of the RGC network.}
\end{figure}

The rest of this paper is organized as follows: Section 2 introduces the related work and the differences between our proposed methods and the related work. Section 3 presents an overview of the proposed RGC network and describes three modules in detail. Section 4 compares the proposed network with several state-of-the-art networks using the nuScenes dataset. Finally, section 5 draws the conclusion.

\section{Related Work}
\label{sec:relate}

In section 2, we will introduce BEV semantic segmentation related work and graph convolutional network related work. In addition, we will briefly discuss the differences between our proposed method and the related works.

\subsection{BEV Semantic Map Segmentation}
BEV semantic map segmentation is divided into single-view segmentation and multi-view segmentation. Since it is challenging to rebuild the whole BEV semantic map from the monocular picture captured by the front-view vehicle camera, earlier single-view segmentation \cite{zhu2021monocular,ammar2019geometric} often uses the high perspective dataset \cite{dosovitskiy2017carla}. With the emergence of multi-view datasets \cite{waymo,caesar2020nuscenes}, current approaches use intrinsic and extrinsic matrices of the camera to translate multi-view autonomous driving datasets into the BEV semantic map.  Two popular categories of segmentation techniques are transformer-based segmentation \cite{cvt, li2022bevformer, xu2022cobevt, pan2020cross} and deep CNNs-based segmentation \cite{LSS, zhang2022beverse, xie2022m2bev, liu2022bevfusion}.

Cross-View Transformer (CVT) \cite{cvt} is one type of the transformer-based segmentation technique. It uses a camera-aware transformer together with intrinsic and extrinsic matrices of cameras to obtain the map-view segmentation from multi-view monocular patches. Like most transformers, CVT focuses attention on the local patch information to extract BEV features rather than the overall region relationship. Cooperative Bird's Eye View Transformer (CoBEVT) \cite{xu2022cobevt} is another type of transformer-based segmentation technique. It uses sparsely sampled attention to obtain sparse global information, which cannot cover the whole global feature map.  In other words, it considers part of the global information rather than the whole global relationship.

Lift, Splat, Shoot (LSS) \cite{LSS} is the pioneer of the deep CNN-based segmentation technique, which obtains a feature map frustum in each view, combines all frustum into a unified BEV intermediate representation using the geometric connection of the cameras, and then obtains map segmentation results. BEVFusion \cite{liu2022bevfusion} extends LSS into a multi-sensor framework to combine the camera and LiDAR BEV feature maps to achieve the best performance with a higher computational cost. However, because of the limited receptive field of each convolutional layer, LSS, BEVFusion, and other CNN-based networks cannot capture the overall region relationship at the early stage of the networks.

\subsection{Graph Convolutional Network}

Graph Convolutional Networks (GCNs) is initially developed for processing graph-structured data. The pioneering work in GCNs \cite{kipf2016semi} introduces a simplified formulation for efficient and scalable graph convolutions, which allows for effective learning on large-scale graph data such as human motion data, coronary arteries data, and so on. Two representative work include MSR-GCN \cite{dang2021msr} and CPR-GCN \cite{yang2020cpr}, where the former has outstanding performance on human motion prediction and the latter outperforms state-of-the-art approaches of automated anatomical labeling. Recently, researchers have explored to apply GCNs to image data by representing images as graphs, to provide new perspectives and avenues. 
Compared with the deep CNNs, GCNs have the benefit of learning the relationships between nodes, which could be defined flexibly. However, the applications of GCNs in image domain \cite{chen2019graph, zhang2019dual, zhang2020dynamic} mainly consider the single image input, which always fail to obtain relational information between multi-view images. 

In this paper, we propose a Residual Graph Convolutional (RGC) module to operate on multi-view images to generate BEV features and obtain a RGC network to improve the performance of the segmentation network.  We apply the RGC module at the early stage of deep CNNs to acquire the global region relationship and reuse the coordinate information to improve the segmentation performance.  Unlike current GCN networks, the RGC network first utilizes multi-view images to generate BEV features. It then employs a non-overlapping graph space projection to efficiently project the complete BEV features, which are regarded as nodes, into the graph space to capture contextual relationships at the early layer of networks. It also utilizes a downsample residual process to enhance coordinate feature reuse and information flow.  In summary, the proposed RGC network is able to simultaneously consider the contextual relationships and the global information to improve the segmentation performance.

\section{Method}
\label{sec:method}

\begin{figure*}[t]
 \includegraphics[width=0.9\linewidth]{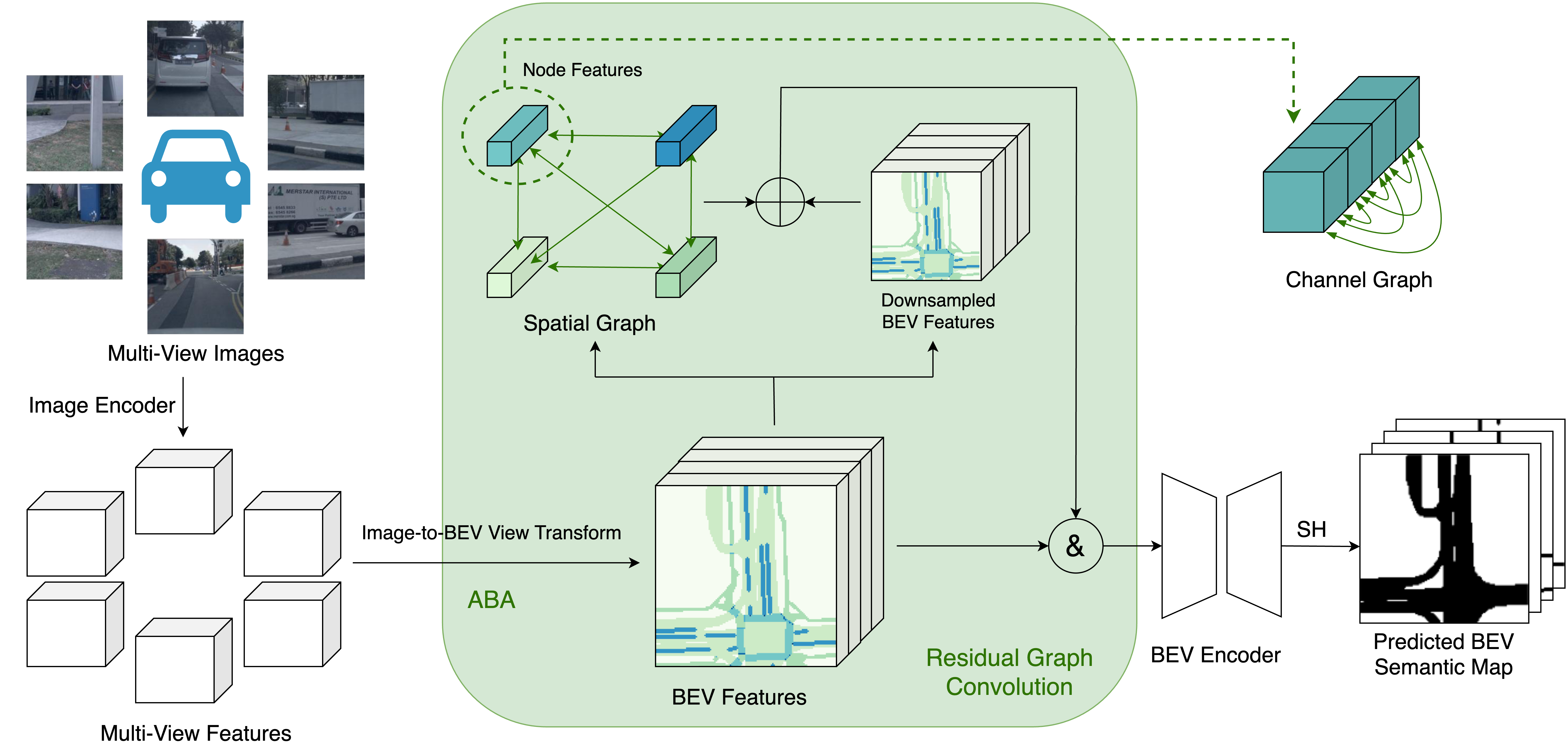}
 \centering
  \caption{Overall architecture of the RGC network.}
\end{figure*}

\begin{figure}[t]
 \includegraphics[width=0.9\linewidth]{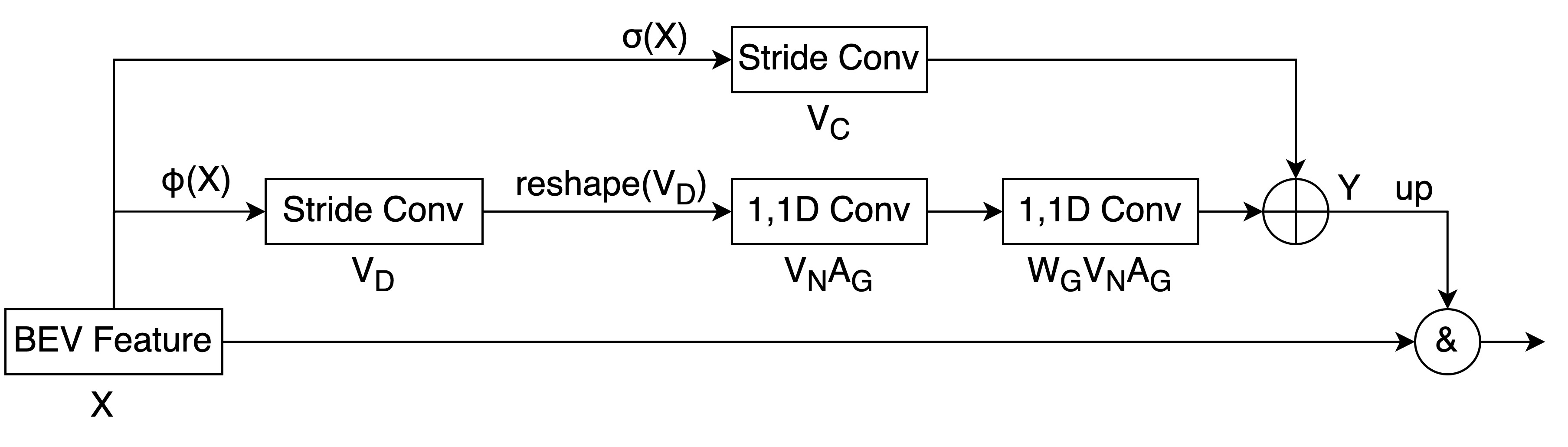}
 \centering
  \caption{Illustration of all operations in the RGC module, where \& represents the concatenation operation.}
\end{figure}

We propose a novel RGC network for BEV semantic segmentation, which utilizes multi-view image inputs to generate graph-based BEV features and produce BEV segmentation results. The RGC network efficiently combines the region interaction information and the BEV coordinate information to increase  
segmentation accuracy. In this section, we first provide an overview of the RGC network. We then briefly introduce each module of the RGC network. Finally, we describe the architecture of 
three proposed modules in detail.

\subsection{Overview}
\label{ssec:overview}
In general, the RGC network could utilize any multi-view BEV network as the backbone. In this paper, we build our own 
modular segmentation network on BEVDet \cite{huang2021bevdet}, a multi-view 3D object detector consisting of a simpler modular structure compared with other networks. In addition to four modules (i.e., image encoder, image-to-BEV view transform, BEV encoder, and 3D object detection head) of BEVDet, we add the segmentation data augmentation and alignment module to enrich the features and inhibit the overfitting problem.  We also add the RGC module to estimate relationships between each region in spatial and channel dimensions while maintaining the coordinate information using its residuals. We further replace the 3D object detection head with the semantic segmentation head to obtain segmentation of multiple regions. The overall architecture of the RGC network is presented in Fig. 2.

\subsection{Functionality of Each Module}
\label{ssec:functionality}
Fig. 3 illustrates all the operations involved in the RGC module. Each module of the RGC network is briefly explained below:

\textbf{Image Encoder:} It extracts multi-scale features of each view image captured by an autonomous vehicle from different horizontal views by utilizing the most recent backbone, SwinTransformer~\cite{liu2021swin}. It then combines features of different resolutions to obtain compact multi-view features via the conventional LSS-based neck \cite{LSS}.

\textbf{Image-to-BEV View Transform:} It \cite{LSS,huang2021bevdet} projects features of each view into coherent BEV features by predicting multi-view depths, combining them to a unified pseudo point cloud, and applying a vertical pooling process. 

\textbf{The Proposed Segmentation Data Augmentation and Alignment:} It augments coherent BEV features, and ground truths by applying common data augmentation techniques (rotation, flipping, and scaling) and simultaneously aligns augmented BEV features with the ground region and 3D object ground truths to enrich features and inhibit overfitting and misalignment issues.

\textbf{The Proposed Residual Graph Convolution:} It uses the spatial graph and the channel graph to extract spatial information between each node and the channel information within each node, respectively.  It utilizes the downsample residual structure to enhance the coordinate information and the non-overlapping graph space projection to project the complete BEV information into the graph space. 

\textbf{BEV Encoder:} 
It uses a backbone and a neck to construct the BEV encoder structure to further process BEV features, which contain graph and coordinate information.

\textbf{The Proposed Semantic Segmentation Head:} It utilizes BEV features to engender the BEV segmentation result for the corresponding category, i.e. drivable area, ped-crossing, walkway, stop-line, carpark-area, and divider.

In the following subsections, we will explain the proposed three modules in more detail. 


\subsection{Three Proposed Modules}
\label{ssec:modules} 

\subsubsection{Segmentation Data Augmentation and Alignment Module}
\label{ssec:project} 
The segmentation data augmentation and alignment module aims to inhibit the overfitting and misalignment problem in BEV domains by augmenting and enriching BEV features and ground truths,  maintaining the semantic consistency (e.g. the 3D vehicles are on the drivable area), and preserving semantic alignment.  Unlike the augmentation in BEVFormer, BEVFusion, CVT, and BEVDet, the proposed module preserves semantic alignment between the augmented object and segmentation ground truths and semantic alignment between augmented BEV map and its augmented ground truths.

We apply some common data augmentation techniques, including rotation, flipping, and scaling, to achieve this goal due to their success in related 3D object detection methods \cite{huang2021bevdet,qi2017pointnet++,yang20203dssd}.  Given $3 \times 3$ rotation and flipping transformer matrices $M_{Rot}$ and $M_{Flip}$ and the scaling transformer parameter $S$, the augmented BEV feature map $B_{AUG-BEV}$ is generated by:
\begin{equation}
  B_{AUG-BEV} = S \times M_{Flip} \times M_{Rot} \times B_{BEV}
  \label{eq:important}
\end{equation}
where $B_{BEV}$ is the BEV feature map. 
In order to automatically align ground truth with augmented BEV feature maps, the corresponding rotation angle is 
computed by the Euler angle formula \cite{eberly2008euler}:
\begin{equation}
  angle = \arctan\left(\frac{M_{Rot}^{-1}(2,1)}{M_{Rot}^{-1}(1,1)}\right)
  \label{eq:angle}
\end{equation}
where $M_{Rot}^{-1}(x,y)$ denotes the element at the coordinate of $(x, y)$ of the inverse rotation matrix.

The augmented ground truth segmentation results $ T_{AUG}$ are estimated by: 
\begin{equation}
  T_{AUG} = S \times M_{Flip} \times T(loc, angle, size)
  \label{eq:GT_map}
\end{equation}
where $T(loc, angle, size)$ is the ground truth segmentation result rotated by $angle$ at location $loc$ with a size of $size$.

\subsubsection{Residual Graph Convolutional (RGC) Module}
The RGC module aims to learn the relationship between each region along both spatial and channel dimensions while maintaining the coordinate information. It consists of three sub-modules: graph space projection, RGC layers, and coordinate space reprojection.

\textbf{Graph Space Projection.} To project BEV coordinate features into the graph space, we utilize the stride convolution operation $\varphi$ to convert input BEV features $X \in \mathbb{R}^{C \times H \times W}$ to downsampled BEV features $V_{D} \in \mathbb{R}^{C \times \frac{H}{d} \times \frac{W}{d}}$: 
$V_{D} = \varphi(X)$ where $d$ represents the stride size.  Since the filters have a kernel size of $ d \times d $ and are not overlapping,
the convolution operation
$\varphi$ is able to efficiently gather the complete BEV information.

\begin{table*}
\small
\centering
\begin{tabular}{llllllll}
\hline
            & drivable area & ped\_crossing & walkway & stop\_line & carpark\_area & divider & mean \\ \hline
OFT \cite{roddick2018orthographic}        & 74.0            & 35.3          & 45.9    & 27.5       & 35.9          & 33.9    & 42.1 \\
LSS \cite{LSS}         & 75.4          & 38.8          & 46.3    & 30.3       & 39.1          & 36.5    & 44.4 \\
CVT \cite{cvt}        & 74.3          & 36.8          & 39.9    & 25.8       & 35.0            & 29.4    & 40.2 \\
BEVFormer \cite{li2022bevformer}   & 80.7          & -          & -    & -       & -          & 21.3    & - \\ 
BEVFusion \cite{liu2022bevfusion}   & 81.7          & 54.8          & 58.4    & 47.4       & 50.7          & 46.4    & 56.6 \\ \hline
\textbf{RGC (ours)} & 
\textbf{81.7}          & \textbf{57.1}          & \textbf{60.5}    & \textbf{51.7}       & \textbf{53.8}          & \textbf{53.5}    & \textbf{59.7} \\ \hline
\end{tabular}
\caption{Comparison of the semantic segmentation results of the proposed RGC network and five state-of-the-art BEV networks. All methods compute the IoU value to evaluate the segmentation performance. The last column denotes the average of IoU values of six classes, where the larger value indicates the better performance. We use bold to highlight the best results for each of six classes and the best overall result.}
\end{table*}

\begin{table*}
\small
\centering
\begin{tabular}{l|ll|lllllll|l}
\hline
variant & augment & graph & drivable area & ped\_crossing & walkway & stop\_line & carpark\_area & divider & mean & FPS \\ \hline
A       &         &       & 67            & 32.8          & 36.9    & 29.1       & 31.9          & 31.2    & 38.2 &\textbf{15.0} \\
B       &         & RGC   & 72.3          & 37.8          & 44.7    & 33.9       & 38.4          & 41.6    & 44.8 &14.7 \\
C       & \checkmark   &       & 80.5          & 54            & 58.2    & 47.7       & 52.3          & 51.3    & 57.3 &14.7 \\
D       & \checkmark   & GloRe  & 80.9          & 54.4          & 58.5    & 48.6       & 51.8          & 51.7    & 57.7 &14.6 \\
E       & \checkmark   & RGC   & \textbf{81.7}          & \textbf{57.1}          & \textbf{60.5}    & \textbf{51.7}       & \textbf{53.8}         & \textbf{53.5}    & \textbf{59.7} &14.5 \\ \hline
\end{tabular}
\caption{
Ablation study of different combinations of data augmentation and graph strategies. Augment denotes the proposed segmentation data augmentation and alignment module. RGC denotes the proposed residual graph convolutional module. GloRe denotes the traditional global reasoning module. All methods compute the IoU and FPS values to evaluate the segmentation performance and  models' inference speed. Higher IoU values indicate  better segmentation results and higher FPS values indicate faster inference speed. We use bold to highlight the best results for each of six classes and the best overall result.
}
\end{table*}

\textbf{RGC Layers.}  
We obtain downsampled BEV features 
$V_{D}$ after the graph space projection. With the reshaping process, we transform $V_{D}$ to graph node features $V_{N} \in \mathbb{R}^{C \times \frac{HW}{d^{2}}}$,  
where $\frac{HW}{d^{2}}$ 
represents the node number $D_{N}$ and $C$ represents the channel number for each node. 
In order to get the relationship between every region in the BEV domain from both the spatial and channel dimensions, we build two completely linked graphs, namely the spatial graph and the channel graph, in the RGC module. The spatial graph learns the relationship between each node, where a node represents multi-channel node features and an edge represents the relationships between node features. The channel graph learns the interdependency between each channel, where a node represents node features along one channel and an edge represents the relationships between channels of each node feature. The two graphs contain the node adjacency matrix $A_{G} \in \mathbb{R}^{D_{N} \times D_{N}}$ and the channel-specific weights $W_{G} \in \mathbb{R}^{C \times C}$ for each node. They are used to compute interactive features $V_{G} \in \mathbb{R}^{C \times \frac{HW}{d^{2}}}$ of the RGC layer by
\begin{equation}
V_{G} = W_{G} \times V_{N} \times A_{G}
\end{equation}

During the training process, $A_{G}$ and $W_{G}$ are randomly initialized and optimized along with other network parameters by the Stochastic Gradient Descent (SGD) method.
It is worth noting that $V_{G}$ has the same dimension as $V_{N}$.  However, it not only captures the relationship of the nodes but also the relationship of the channels within each node. We further reshape $V_{G}$ to $Y_{G}$ to have the same dimension as $V_{D}$. To seamlessly combine the low-resolution coordinate features and graph features, we propose a downsample residual process to obtain combined features $Y$ by
\begin{equation}
Y = Y_{G} + \sigma (X)
\end{equation}
where $\sigma$ represents a downsampling convolution operation to transform $X$ to $V_{C} \in \mathbb{R}^{\frac{C \times H \times W}{d^{2}}}$ with the same dimension of $Y_{G}$.

\textbf{Coordinate Space Reprojection.} After the graph interaction, we utilize the upsampling operation to reproject $Y$ back to the original BEV space $\mathbb{R}^{C \times H \times W}$ to be consistent with the network architecture.
Specifically, the bilinear interpolation $F_{bilinear}$ is adopted to upsample $Y$ by $d$ times to obtain residual graph features $X_{G}$: 
\begin{equation}
X_{G} = F_{bilinear}(Y,scale=d)
\end{equation}
After reprojection, the residual graph features $X_{G}$ are finally concatenated with input BEV features $X$ to maintain 
both original information and processed information inputs as 
($Concat(X,X_{G})$). 

\subsubsection{Semantic Segmentation Head Module}
We utilize nine convolutional layers including eight $3 \times 3$ and one $1 \times 1$ convolutional layers to get the binary mask for each category.

\section{Experimental Results}
\label{sec:pagestyle}

\begin{figure*}[t]
 \includegraphics[width=0.9\linewidth]{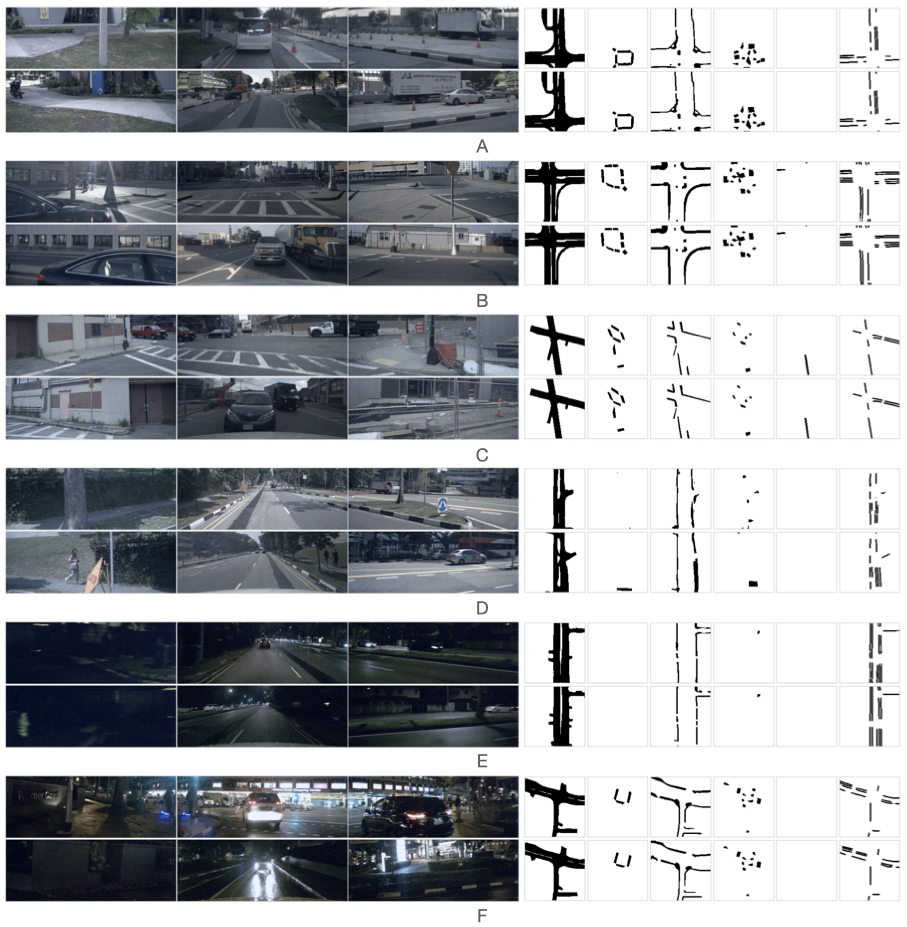}
 \centering
  \caption{Illustration of qualitative segmentation results of six sets of multiple view input images. The six view images in each set surround the ego-vehicle and are displayed on the left, where the top three images are front view images and the bottom three images are back view images. The upper row on the right presents our predicted BEV segmentation for drivable area, ped-crossing, walkway, stop-line, carpark-area, and dividers. The lower row on the right presents the ground truth segmentation for the same six classes for easy comparison. The qualitative segmentation results show that the proposed method is able to accurately segment wide roads, but it does not sense the narrow paths well.
  }
  \label{fig:qualitative_results}
\end{figure*}

We utilize the benchmark dataset nuScenes \cite{caesar2020nuscenes} to evaluate the proposed RGC network and all compared networks. The nuScenes dataset consists of 1,000 driving scenes collected in Boston Seaport and Singapore’s One North, Queenstown and Holland Village districts. It is one of the most popular datasets for semantic segmentation in autonomous vehicle. It consists of 700 scenes in the training set, 150 scenes in the validation set, and 150 scenes in the testing set. Every scene contains a LiDAR set and colorful images from six view cameras. Each scene is also labeled with semantic mask annotations for 11 semantic classes and additional bitmaps. 
In our experiment, we choose six important segmentation classes for evaluation, which are also commonly used in other compared state-of-the-art networks \cite{liu2022bevfusion}.  These semantic classes include drivable area, ped-crossing, walkway, stop-line, carpark-area, and dividers.

\subsection{Implementation}
\label{ssec:Implementations}
The overview of RGCN is shown in Fig. 2. 
To maintain high computational efficiency, all multi-view input images are downsampled to $256 \times 704$ in all experiments.  
Multi-view image features are then extracted using the up-to-date network SwinTransformer \cite{liu2021swin}, which is pre-trained on the ImageNet \cite{krizhevsky2017imagenet}. They are further processed by the image-to-BEV view transform to get unified BEV features, which have 64 channels and a $128 \times 128$ resolution. After incorporating the proposed segmentation data augmentation and alignment and RGC modules, the channel number of  BEV features is doubled after the concatenation operation. At last, the BEV encoder utilizes ResNet to further process BEV features. The proposed semantic segmentation head module generates the segmentation mask for each semantic class.

\subsection{Quantitative Results}
\label{ssec:Results}
We utilize two metrics, namely, Intersection-over-Union (IoU) and mean IoU (mIoU), to evaluate the performance of all networks. Table 1 compares the performance of the proposed RGC network with five state-of-the-art BEV segmentation networks \cite{ roddick2018orthographic, LSS, cvt, liu2022bevfusion, li2022bevformer} on six categories in terms of IoU and mIoU, which were reproduced by \cite{liu2022bevfusion}. To ensure a fair comparison, we use the results of the single timestamp BEVFormer model.  BEVFormer \cite{li2022bevformer} reports a drivable area accuracy of 80.7\% and a divider accuracy of 21.3\%, which are respectively lower than the accuracy of RGC by 1.0\% and 32.2\%. Each BEV query of BEVFormer \cite{li2022bevformer} only interacts with image features in the regions of interest. On the contrary, the proposed RGC model considers all regions of BEV features to more thoroughly consider global semantic relationship to achieve better performance as shown in Table 1. In general, the RGC network outperforms the other four segmentation networks by at least 3.1\% in terms of
mIoU. Specifically, the RGC network achieves a higher IoU of 2.3\%, 2.1\%, 4.3\%, 3.1\%, and 7.1\% than the best-compared network (i.e., BEVFusion) for five semantic classes of ped-crossing, walkway, stop-line, carpark-area, and divider, respectively. It performs similarly as BEVFusion on the drivable area containing a large region. The RGC network achieves a higher IoU of at least 6.3\%, 18.3\%, 14.2\%, 21.4\%, 14.7\%, and 17.0\% than the other three compared networks for six semantic classes of drivable area, ped-crossing, walkway, stop-line, carpark-area, and divider, respectively.  In summary, the RGC network outperforms the four compared networks in all six semantic classes, especially in five classes with small and delicate regions. 

\subsection{Ablation Study}
\label{ssec:Ablation Study}
Table 2 compares the performance of the proposed RGC network and four variants in terms of IoU in each category and mIoU.  Variant A is the baseline network without segmentation data augmentation and alignment and RGC modules. Variants B and C are the baseline network adding the RGC module and the segmentation data augmentation and alignment module, respectively. Variant D is variant C adding a GloRe \cite{chen2019graph} module, which has similar functionality as the RGC module. Other graph convolutional networks such as MSR-GCN \cite{dang2021msr} and CPR-GCN \cite{yang2020cpr} cannot process the image dataset. As a result, we do not include their variants to compare with the performance of the proposed RGC module. Variant E is the proposed RGC network (i.e., variant C adding the RGC module). Table 2 shows that variant B improves the baseline network by 6.6\% in mIoU due to the incorporation of RGC and variant C improves the baseline network by 19.1\% in mIoU due to the incorporation of segmentation data augmentation and alignment. Variant D improves the accuracy of all semantic classes except for the carpark-area when compared with variant C.  Variant E (i.e., the proposed RGC module) significantly improves the segmentation accuracy of all semantic classes when compared with variant C. It also outperforms variant D with a higher mIoU of 2.0\% due to the incorporation of RGC instead of GloRe.  We believe this improvement may mainly attribute to: 1) extracted spatial information between each node and channel information within each node resulting from the two interconnected spatial and channel graphs; 2) the enhanced coordinate feature reuse resulting from the downsample residual process in RGC layers;  
3) the similar semantic regions clustered and represented by a node in the graph region; 4) the usefully connected relationships between different BEV-space feature dimensions after reasoning the relationships of different regions; and 5) the projection of the connected information to the coordinate space. Specifically, the interconnected graphs extract node features from BEV features to describe relationships between regions in terms of both spatial and channel dimensions to improve the segmentation accuracy.  In addition, the downsample residual process enhances the coordinate feature space to maintain both coordinate and contextual relationship information to efficiently estimate global contextual relationships to improve the segmentation accuracy. It should be mentioned that adding the RGC module or the data augmentation module makes the inference speed of variant B or variant C slower than variant A by 0.3 FPS. Simultaneously adding RGC and data augmentation modules makes the inference speed of variant E slower than variant A by 0.5 FPS and the accuracy of variant E higher than variant A by 21.5\%. In general, adding proposed modules leads to a little slower computational time and significantly higher accuracy (i.e., mIoU).

\subsection{Qualitative Results}
\label{ssec:Qualitative Results}

\begin{figure}[t]
 \includegraphics[width=1\linewidth]{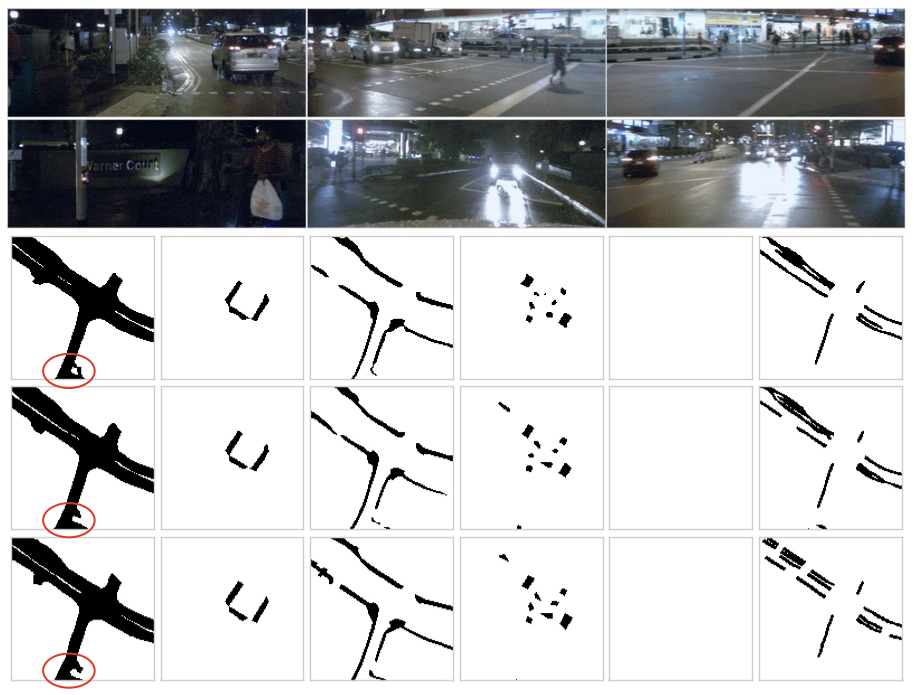}
 \centering
  \caption{Illustration of one sample scene of six views (first two rows), segmentation result of Variant C (third row), segmentation results of Variant E (fourth row), and the ground-truth (fifth row). The inaccurate segmentation results are circled in red.}
  \label{fig:qualitative}
\end{figure}

Figure 4 presents six sample scenes, their segmentation results obtained by the proposed RGC network, and their ground-truth segmentation results. Figures 4(A),(B),(C), and (D) are scene shots in the daytime, while Figure 4(E) and (F) are scene shots at night. Figures 4(E) shows the segmentation results are not perfect due to the lack of light at night. However, we observe that Figure 4(F) shows that the segmentation results of the RGC network on a night scene are satisfactory when there is sufficient light at night. In general, the RGC network is able to produce segmentation results that match well with ground truths.

Figure 5 presents the qualitative results of a traditional deep CNN-based segmentation network without an RGC module (Variant C on the third row of Table 2) and the proposed RGC network (Variant E on the fifth row of Table 2) on one sample scene. We observe that the traditional deep CNN-based segmentation network cannot accurately segment the border of a delicate region as circled in red and the RGC network produces more accurate segmentation results due to its accurate estimation of global contextual relationships.

\section{Conclusion}
\label{sec:typestyle}
In this paper, we propose a novel network that employs an RGC module to estimate global contextual relationships via interconnected spatial and channel graphs and maintain the enhanced coordinate information via a downsample residual process. A non-overlapping graph space projection is utilized to efficiently project the complete BEV information into graph space. A segmentation data augmentation and alignment module is further incorporated to enrich features and align BEV features and ground truth to geometrically preserve their semantic alignment and inhibit the overfitting and misalignment issue. The RGC network obtains state-of-the-art performance on the nuScenes benchmark dataset when compared with four state-of-the-art networks and its four variants in terms of both IoU and mIoU.

{\small
\bibliographystyle{ieee_fullname}
\bibliography{main.bbl}

\begin{thebibliography}{10}\itemsep=-1pt

\bibitem{ammar2019geometric}
Syed Ammar~Abbas and Andrew Zisserman.
\newblock A geometric approach to obtain a bird's eye view from an image.
\newblock In {\em Proceedings of the IEEE/CVF International Conference on Computer Vision Workshops}, pages 0--0, 2019.

\bibitem{caesar2020nuscenes}
Holger Caesar, Varun Bankiti, Alex~H Lang, Sourabh Vora, Venice~Erin Liong, Qiang Xu, Anush Krishnan, Yu Pan, Giancarlo Baldan, and Oscar Beijbom.
\newblock nuscenes: A multimodal dataset for autonomous driving.
\newblock In {\em Proceedings of the IEEE/CVF conference on computer vision and pattern recognition}, pages 11621--11631, 2020.

\bibitem{chen2019graph}
Yunpeng Chen, Marcus Rohrbach, Zhicheng Yan, Yan Shuicheng, Jiashi Feng, and Yannis Kalantidis.
\newblock Graph-based global reasoning networks.
\newblock In {\em Proceedings of the IEEE/CVF Conference on Computer Vision and Pattern Recognition}, pages 433--442, 2019.

\bibitem{dang2021msr}
Lingwei Dang, Yongwei Nie, Chengjiang Long, Qing Zhang, and Guiqing Li.
\newblock Msr-gcn: Multi-scale residual graph convolution networks for human motion prediction.
\newblock In {\em Proceedings of the IEEE/CVF International Conference on Computer Vision}, pages 11467--11476, 2021.

\bibitem{dosovitskiy2017carla}
Alexey Dosovitskiy, German Ros, Felipe Codevilla, Antonio Lopez, and Vladlen Koltun.
\newblock Carla: An open urban driving simulator.
\newblock In {\em Conference on robot learning}, pages 1--16. PMLR, 2017.

\bibitem{eberly2008euler}
David Eberly.
\newblock Euler angle formulas.
\newblock {\em Geometric Tools, LLC, Technical Report}, pages 1--18, 2008.

\bibitem{huang2021bevdet}
Junjie Huang, Guan Huang, Zheng Zhu, and Dalong Du.
\newblock Bevdet: High-performance multi-camera 3d object detection in bird-eye-view.
\newblock {\em arXiv preprint arXiv:2112.11790}, 2021.

\bibitem{kipf2016semi}
Thomas~N Kipf and Max Welling.
\newblock Semi-supervised classification with graph convolutional networks.
\newblock {\em arXiv preprint arXiv:1609.02907}, 2016.

\bibitem{krizhevsky2017imagenet}
Alex Krizhevsky, Ilya Sutskever, and Geoffrey~E Hinton.
\newblock Imagenet classification with deep convolutional neural networks.
\newblock {\em Communications of the ACM}, 60(6):84--90, 2017.

\bibitem{li2022bevformer}
Zhiqi Li, Wenhai Wang, Hongyang Li, Enze Xie, Chonghao Sima, Tong Lu, Qiao Yu, and Jifeng Dai.
\newblock Bevformer: Learning bird's-eye-view representation from multi-camera images via spatiotemporal transformers.
\newblock {\em arXiv preprint arXiv:2203.17270}, 2022.

\bibitem{liu2021swin}
Ze Liu, Yutong Lin, Yue Cao, Han Hu, Yixuan Wei, Zheng Zhang, Stephen Lin, and Baining Guo.
\newblock Swin transformer: Hierarchical vision transformer using shifted windows.
\newblock In {\em Proceedings of the IEEE/CVF International Conference on Computer Vision}, pages 10012--10022, 2021.

\bibitem{liu2022bevfusion}
Zhijian Liu, Haotian Tang, Alexander Amini, Xinyu Yang, Huizi Mao, Daniela Rus, and Song Han.
\newblock Bevfusion: Multi-task multi-sensor fusion with unified bird's-eye view representation.
\newblock {\em arXiv preprint arXiv:2205.13542}, 2022.

\bibitem{pan2020cross}
Bowen Pan, Jiankai Sun, Ho~Yin~Tiga Leung, Alex Andonian, and Bolei Zhou.
\newblock Cross-view semantic segmentation for sensing surroundings.
\newblock {\em IEEE Robotics and Automation Letters}, 5(3):4867--4873, 2020.

\bibitem{LSS}
Jonah Philion and Sanja Fidler.
\newblock Lift, splat, shoot: Encoding images from arbitrary camera rigs by implicitly unprojecting to 3d.
\newblock In {\em European Conference on Computer Vision}, pages 194--210. Springer, 2020.

\bibitem{qi2017pointnet++}
Charles~Ruizhongtai Qi, Li Yi, Hao Su, and Leonidas~J Guibas.
\newblock Pointnet++: Deep hierarchical feature learning on point sets in a metric space.
\newblock {\em Advances in neural information processing systems}, 30, 2017.

\bibitem{roddick2018orthographic}
Thomas Roddick, Alex Kendall, and Roberto Cipolla.
\newblock Orthographic feature transform for monocular 3d object detection.
\newblock {\em arXiv preprint arXiv:1811.08188}, 2018.

\bibitem{waymo}
Pei Sun, Henrik Kretzschmar, Xerxes Dotiwalla, Aurelien Chouard, Vijaysai Patnaik, Paul Tsui, James Guo, Yin Zhou, Yuning Chai, Benjamin Caine, et~al.
\newblock Scalability in perception for autonomous driving: Waymo open dataset.
\newblock In {\em Proceedings of the IEEE/CVF conference on computer vision and pattern recognition}, pages 2446--2454, 2020.

\bibitem{xie2022m2bev}
Enze Xie, Zhiding Yu, Daquan Zhou, Jonah Philion, Anima Anandkumar, Sanja Fidler, Ping Luo, and Jose~M Alvarez.
\newblock M\^{} 2bev: Multi-camera joint 3d detection and segmentation with unified birds-eye view representation.
\newblock {\em arXiv preprint arXiv:2204.05088}, 2022.

\bibitem{xu2022cobevt}
Runsheng Xu, Zhengzhong Tu, Hao Xiang, Wei Shao, Bolei Zhou, and Jiaqi Ma.
\newblock Cobevt: Cooperative bird's eye view semantic segmentation with sparse transformers.
\newblock {\em arXiv preprint arXiv:2207.02202}, 2022.

\bibitem{yang2020cpr}
Han Yang, Xingjian Zhen, Ying Chi, Lei Zhang, and Xian-Sheng Hua.
\newblock Cpr-gcn: Conditional partial-residual graph convolutional network in automated anatomical labeling of coronary arteries.
\newblock In {\em Proceedings of the IEEE/CVF conference on computer vision and pattern recognition}, pages 3803--3811, 2020.

\bibitem{yang20203dssd}
Zetong Yang, Yanan Sun, Shu Liu, and Jiaya Jia.
\newblock 3dssd: Point-based 3d single stage object detector.
\newblock In {\em Proceedings of the IEEE/CVF conference on computer vision and pattern recognition}, pages 11040--11048, 2020.

\bibitem{zhang2019dual}
Li Zhang, Xiangtai Li, Anurag Arnab, Kuiyuan Yang, Yunhai Tong, and Philip~HS Torr.
\newblock Dual graph convolutional network for semantic segmentation.
\newblock {\em arXiv preprint arXiv:1909.06121}, 2019.

\bibitem{zhang2020dynamic}
Li Zhang, Dan Xu, Anurag Arnab, and Philip~HS Torr.
\newblock Dynamic graph message passing networks.
\newblock In {\em Proceedings of the IEEE/CVF Conference on Computer Vision and Pattern Recognition}, pages 3726--3735, 2020.

\bibitem{zhang2022beverse}
Yunpeng Zhang, Zheng Zhu, Wenzhao Zheng, Junjie Huang, Guan Huang, Jie Zhou, and Jiwen Lu.
\newblock Beverse: Unified perception and prediction in birds-eye-view for vision-centric autonomous driving.
\newblock {\em arXiv preprint arXiv:2205.09743}, 2022.

\bibitem{cvt}
Brady Zhou and Philipp Kr{\"a}henb{\"u}hl.
\newblock Cross-view transformers for real-time map-view semantic segmentation.
\newblock In {\em Proceedings of the IEEE/CVF Conference on Computer Vision and Pattern Recognition}, pages 13760--13769, 2022.

\bibitem{zhu2021monocular}
Minghan Zhu, Songan Zhang, Yuanxin Zhong, Pingping Lu, Huei Peng, and John Lenneman.
\newblock Monocular 3d vehicle detection using uncalibrated traffic cameras through homography.
\newblock In {\em 2021 IEEE/RSJ International Conference on Intelligent Robots and Systems (IROS)}, pages 3814--3821. IEEE, 2021.

\end{thebibliography}
}

\end{document}